\documentclass[10pt,twocolumn,letterpaper]{article}

\usepackage{cvpr}              

\usepackage{graphicx}
\usepackage{amsmath}
\usepackage{amssymb}
\usepackage{booktabs}
\usepackage{color}
\usepackage[table,xcdraw]{xcolor}
\usepackage{multirow}
\usepackage{algorithm}
\usepackage{algorithmic}

\usepackage[pagebackref,breaklinks,colorlinks]{hyperref}

\usepackage[capitalize]{cleveref}
\crefname{section}{Sec.}{Secs.}
\Crefname{section}{Section}{Sections}
\Crefname{table}{Table}{Tables}
\crefname{table}{Tab.}{Tabs.}

\def\cvprPaperID{99} 
\def\confName{CVPR}
\def\confYear{2023}

\begin{document}

\title{SwinFSR: Stereo Image Super-Resolution using SwinIR and Frequency Domain Knowledge}


\author{Ke Chen, Liangyan Li, Huan Liu, Yunzhe Li, Congling Tang and Jun Chen\\
McMaster University, Hamilton, Canada\\
{\tt\small \{chenk59, lil61, liuh127, liy366, tangc61, chenjun\}@mcmaster.ca}}
\maketitle

\begin{abstract}
Stereo Image Super-Resolution (stereoSR) has attracted significant attention in recent years due to the extensive deployment of dual cameras in mobile phones, autonomous vehicles and robots. In this work, we propose a new StereoSR method, named SwinFSR, based on an extension of SwinIR, originally designed for single image restoration, and the frequency domain knowledge obtained by the Fast Fourier Convolution (FFC).
Specifically, to effectively gather global information, we modify the Residual Swin Transformer blocks (RSTBs) in SwinIR by explicitly incorporating the frequency domain knowledge using the FFC and employing the resulting residual Swin Fourier Transformer blocks
(RSFTBs) for feature extraction.
Besides, for the efficient and accurate fusion of stereo views, we propose a new cross-attention module referred to as RCAM,
which achieves highly competitive performance while requiring less computational cost than the state-of-the-art cross-attention modules.
Extensive experimental results and ablation studies demonstrate the effectiveness and efficiency of our proposed SwinFSR.

\end{abstract}

\section{Introduction}
Stereo image pairs can encode 3D scene cues into stereo correspondences between the left and right images. With the extensive deployment of dual cameras in mobile phones, autonomous vehicles and robots, the stereo vision has attracted increasing attention in both academia and industry. In many applications such as AR/VR \cite{li2022super, spagnolo2023design} and robot navigation \cite{okarma2015application}, increasing the resolution of stereo images is highly demanded to attain superior perceptual quality and optimize performance for downstream tasks \cite{wang2022ntire}. Recently, many deep-learning-based methods  \cite{wang2019learning, wang2021symmetric, liang2021swinir, chu2022nafssr} have been proposed to address the stereo super-resolution (stereoSR) problem.

\begin{figure}[hbtp]
\centering
\includegraphics[width=\linewidth]{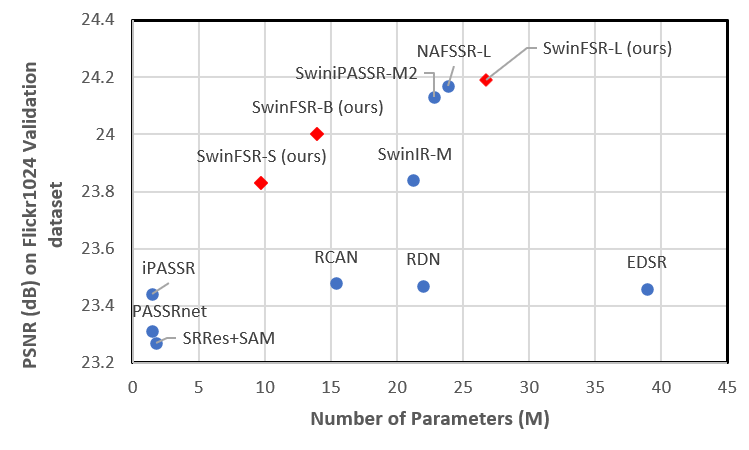}
\caption{Parameters vs. PSNR of models for $4\times$ stereo SR on Flickr1024 \cite{wang2019flickr1024} test set. Our SwinFSR families achieve the highest performance.}
\label{fig10}
\end{figure}

In favour of the remarkable capability of the Transformer \cite{vaswani2017attention},
most recent stereoSR methods \cite{vaswani2017attention,wang2022ntire} are developed based on Transformer structure, especially on a variant for image restoration task, i.e., SwinIR \cite{liang2021swinir}.
However, there are some common issues with
the existing SwinIR based models such as SwiniPASSR \cite{jin2022} and SwinFIR \cite{zhang2022swinfir}. First, SwiniPASSR 
does not have a specifically designed mechanism for exploiting features extracted from two views 
as biPAM \cite{wang2021symmetric} is used by default. 
Second, it focuses on spatial features but not spectral features,  thus failing to make full use of large receptive fields to gather global information in a more direct manner. As of SwinFIR \cite{zhang2022swinfir}, it also does not explicitly exploit the interdependence of features extracted from two views due to a lack of cross attention modules. Moreover, SwinFIR cannot estimate epipolar stereo disparity as it requires squared images as inputs. 

Inspired by the observation of \cite{suvorov2021naejin} regarding the effectiveness of the Fast Fourier Convolution (FFC) block in capturing global information, we modify Residual Swin Transformer blocks (RSTBs) in SwinIR by explicitly incorporating the frequency domain knowledge and employ the resulting Residual Swin Fourier Transformer blocks (RSFTBs) for feature extraction. 
Besides the proposed feature extractor, we also aim to enhance the cross-attention module for effective and efficient informant exchange between two views. 
Instead of directly using the off-the-shelf cross-attention modules such as
SAM \cite{ying2020stereo}, SCAM \cite{chu2022nafssr}, and biPAM \cite{wang2021symmetric},  we propose a new cross-attention module named RCAM. Specifically, to balance between efficient inference and accurate learning, we modify the biPAM by removing the need to handle occlusion and redesigning the attention mechanism. 
Moreover, to address the inflexibility of squared training patches with respect to the epipolar disparity, we modify the local window in the Swin Transformer so that the network can process rectangular input patches. Based on the above innovations, we develop a new stereoSR network, namely SwinFSR.
In summary, our SwinFSR has two branches built with RSFTBs to process left and right views, respectively. The two branches share the same weights. RCAMs are inserted between the two branches to exchange and consolidate cross-view information. 
     
Furthermore, various training/testing strategies are adopted to unleash the potential of SwinFSR. In training, we use several effective data augmentation methods to boost SR performance, such as random cropping, flipping, and channel shuffling. We also conduct experiments to find the best possible hyper-parameters, such as dropout rate \cite{kong2022reflash}, window size, and stochastic depth \cite{huang2016deep} of the Swin Transformer based models. As shown in Figure \ref{fig10}, our SwinFSR families have better performance-complexity trade-offs than the existing methods.

Our contributions can be summarized as follows: 
\begin{itemize}
\item Based on a systematic analysis of the issues with the existing methods, we propose a new stereoSR method, SwinFSR. It inherits the advantages of SwinIR and Fast Fourier Convolution and exploits both spatial and spectral features.

\item We propose a new cross-attention module, named RCAM, that strikes a good balance between efficient inference and accurate learning. This is realized by modifying biPAM to circumvent occlusion handling as well as redesigning its attention mechanism. It is shown that this modification can help expedite the inference speed without significantly jeopardizing the performance. 

\item 
Extensive experimental results demonstrate the effectiveness and efficiency of our proposed approach.
\end{itemize} 
\section{Related Works}
\subsection{Single Image Super-resolution}
Single image super-resolution (SISR) aims to generate high-resolution images based on their low-resolution counterparts. SISR has been extensively researched in the fields of image processing and computer vision, and various approaches have been proposed to address this problem. Super-Resolution Convolutional Neural Networks (SRCNN) \cite{dong2014image} make the first attempt to bring deep learning to bear upon SISR, and subsequent methods VDSR and EDSR \cite{lim2017enhanced, zhang2018image} further take advantage of residual and dense connections to achieve improved performances. Attention mechanisms, including channel attention \cite{zhang2018residual, dai2019second, magid2021dynamic} and channel-spatial attention \cite{liang2021swinir, dai2019image, niu2020single}, have also been proposed as an effective tool for tackling SISR. Recently, in view of its remarkable ability  in natural language processing (NLP), SwinIR \cite{liang2021swinir}, a Transformer-based structure has been employed for SISR, achieving state-of-the-art (SOTA) performance.  

\subsection{Stereo Image Super-Resolution}
Stereo image super-resolution (stereoSR) is a challenging task in computer vision that requires generating high-resolution images from stereo image pairs. Convolutional neural networks (CNNs) are commonly used in deep learning-based stereoSR approaches, such as the Single Image Stereo Matching network (SSRN)  \cite{luo2018single}. It introduces a stereo matching module to establish dense correspondence between low-resolution stereo images and then applies a CNN to enhance the resolution of each image. Attention mechanisms have also been explored in recent works to improve stereoSR. For instance, \cite{wang2019learning} proposes a parallax attention module (PAM) and builds a PASSRnet for stereoSR to handle varying parallax. \cite{zhao2020exploring} designs an attention-based method that can adaptively weigh the stereo features to enhance the resolution of the stereo images. \cite{ying2020stereo} introduces stereo attention modules (SAMs) into pre-trained single image SR (SISR) networks to handle information assimilation.  \cite{song2020stereoscopic} addresses the occlusion issue by using disparity maps regressed by parallax attention maps to assess stereo consistency. \cite{wang2021symmetric} develops an iPASSRnet that uses symmetry cues and a Siamese network equipped with a biPAM structure to super-resolve both left and right images. And NAFSSR \cite{chu2022nafssr}, the winner of the NTIRE 2022 StereoSR Challenge \cite{wang2022ntire}, achieves the SOTA by inserting cross-view attention modules (SCAMs) between consecutive NAFblocks \cite{chen2022simple}. These works have made significant contributions to the stereoSR and have opened up new possibilities for future research in this area. 

In this work, we move one step further by introducing a residual stereo cross-attention module (RCAM). In contrast to SAM \cite{ying2020stereo}, which requires calculating an occlusion map, our RCAM presents a better solution with high efficiency.

\subsection{Vision Transformer}
As a recent advance in the field of computer vision, visual Transformers \cite{vaswani2017attention} have garnered significant attention for their ability to capture long-range dependencies in images, especially for high-level vision tasks such as image classification \cite{dosovitskiy2020image, liu2021swin} and object detection\cite{liu2021swin, wu2020visual, cao2021swin}. Moreover, Transformers have also been applied to low-level vision tasks (see, e.g., \cite{yang2020learning}). To reduce the computational complexity of self-attention operations in Transformers, a hierarchical visual Transformer called  Swin Transformer \cite{cao2021swin} is proposed using shifted window techniques, which achieved SOTA performance on various tasks such as image recognition, object detection, and segmentation. SwinIR \cite{liang2021swinir} and Swin V2 \cite{liu2022swin} have implemented some further refinements to make Transformers more efficient. Overall, these works have demonstrated the effectiveness of visual Transformers in a wide range of computer vision tasks.

\begin{figure*}[!t]
    \centering
    \includegraphics[width=0.9\linewidth]{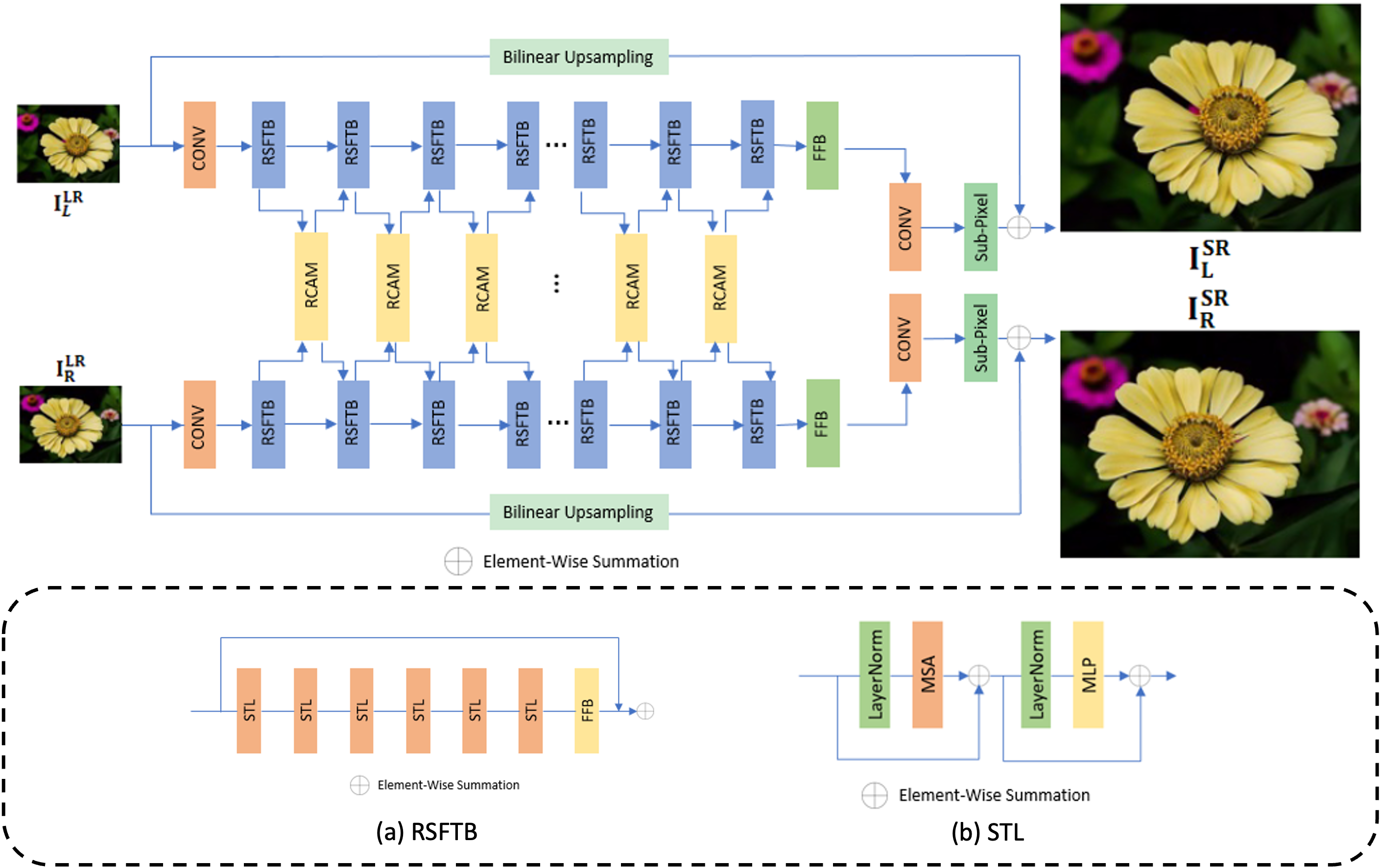}
    \caption{Top: illustration of the proposed SwinFSR Architecture. Bottom: (a) Residual Swin Fourier Transformer Block (RSFTB), (b) Swin Transformer Layer (STL). }
    \label{fig1}
\end{figure*}



\subsection{Training and Testing Strategies}
Regularization methods such as dropout \cite{kong2022reflash} and stochastic depth \cite{huang2016deep} are widely employed to enhance the model performance  in high-level computer vision tasks. Recently, the above regularization methods have been  introduced in image restoration tasks. For example, stochastic depth is employed in \cite{chu2022nafssr}  to address the issue of overfitting to the stereo-training data and improve generalization. Similarly, \cite{kong2022reflash} adjusts the dropout method in their SR tasks. In this work, we will systematically study how the factors such as dropout rate, window size, and stochastic depth can impact PSNR performance in Swin Transformer-based models.  Additionally, since test-time augmentation (TTA) \cite{wang2019automatic, kim2020learning} is a technique that is frequently used in computer vision competitions to boost performance, we also investigate its capability in the context of stereoSR through an ablation study. 

\section{Method}
\noindent In this section, we introduce our method in detail. In Sec \ref{network}, we first give an overview of the network's architecture. In Sec \ref{Strategies}, we then present the training and testing strategies.

\subsection{Network Architecture} \label{network}

\subsubsection{Overall Framework}

Figure \ref{fig1} depicts an outline of our proposed transformer-based Stereo SR network (SwinFSR). SwinFSR takes a low-resolution stereo image pair as input and enhances the resolution of both left and right view images. To be specific, Our SwinFSR has two branches built with RSFTB to process left and right views, respectively. RCAMs described in Figure \ref{fig5}, are inserted between the left and right branches to interact with cross-view information. In essence, SwinFSR is composed of three parts: intra-view feature extraction, cross-view feature fusion, and reconstruction

\noindent
\textbf{Intra-view feature extraction and reconstruction.} To start, a $3 \times 3$ convolutional layer is employed to extract the shallow features from input images. Then, RSFTBlocks are stacked to achieve deep intra-view feature extraction. We will detail the RSFTBlock in Section \ref{rsftb}. Once feature extraction is completed, a Fast Fourier Block (FFB) is applied, followed by a pixel shuffle layer \cite{shi2016real} that upsamples the feature by a scale factor of 4. Additionally, to alleviate the burden of feature extraction, we follow \cite{liang2022vrt, chu2022nafssr} to predict the difference between the bilinearly upsampled low-resolution image and the high-resolution ground truth.

\noindent
\textbf{Cross-view feature fusion.} To engage with information from different views, we incorporate RCAM following every RSFTBlocks. RCAM utilizes stereo features produced by the preceding RSFTBlocks as inputs for conducting bidirectional cross-view interactions and produces interacted features fused with input features from the same view. The details of the RCAM are elaborated in Section \ref{rcam}.

\subsubsection{RSFTBlock.} \label{rsftb}
As shown in Figure \ref{fig1} (a), the residual Swin Transformer block (RSTB) is a residual block built using Swin Transformer Layers (STL) in Figure \ref{fig1} (b) and a Fast Fourier Convolution Block in Figure \ref{fig4}. Given the input feature  $F_{i, 0}$ of the i-th RSFTB, we first extract intermediate features $F_{i, j}$ by L STLs as: 
\begin{equation}
        F_{i,j} = STL_{i,j}(F_{i,j-1}),j = 1,2,3,...,L ,
\end{equation}
\noindent where $STL_{i,j}$ is j-th STL in the i-th RSFTB.

We then feed the feature from L-th STL to FFB to extract frequency domain knowledge. After that, we output the summation of FFB outputs and input features by:
\begin{equation}
        F_{i,out} = FFB_i(F_{i,L}) + F_{i,0},   
\end{equation}

\noindent
 where $FFB_i$ represents the last FFB block in the i-th RSFTB block. And $F_{i,out}$ is the output feature of i-th RSFTB block.

\subsubsection{STL Blocks.} As shown in Figure \ref{fig1} (b), a two-layer multi-layer perceptron (MLP) with fully connected layers and GELU non-linearity between them is used. Prior to using the MSA and MLP, a LayerNorm (LN) layer is attached  and a residual connection is employed for both modules. The complete process for the STL block is explained in detail in \cite{liang2021swinir}.

\subsubsection{Fast Fourier Convolution Blocks (FFB).} 
\begin{figure}[hbtp]
\centering
\includegraphics[width=\linewidth]{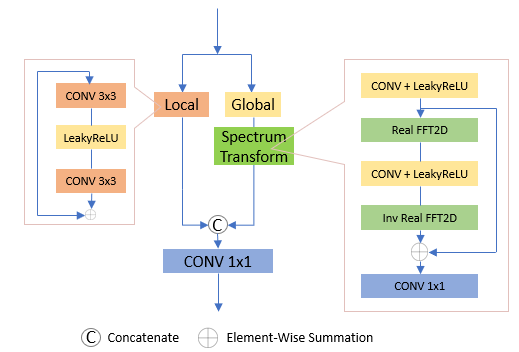}
\caption{Fast Fourier Convolution Block (FFB).}
\label{fig4}
\end{figure}

Our backbone model SwinIR is mainly composed of residual Swin Transformer blocks (RSTBs) that utilize several Swin Transformer layers to achieve local attention and cross-window interaction. However, in the context of stereo SR, it is advantageous to incorporate both local and global information \cite{gu2021interpreting}. To achieve this, we take inspiration from the Fast Fourier Convolution (FFC) \cite{chi2020fast}, which can use the global context in early layers \cite{suvorov2021naejin}. To this end, we propose a hybrid module including an FFC and a residual module called the Fast Fourier Block (FFB) to enhance the model's  ability. 
As shown in Figure \ref{fig4}, the FFB has two main components: a local spatial conventional convolution operation on the left and a global FFC spectrum transform on the right. The outputs from both operations are concatenated and then subjected to a convolution operation to generate the final result. Here we formalize the operation. Given an input feature of FFB block $F_{i, L}$, 
we send $F_{i, L}$ into two distinct branches, local and global. In the local branch, $H_{local}$ is utilized and extracts the local features in the spatial domain, and $H_{global}$ is intended to capture the long-range context in the frequency domain. To increase readability, we use $F$ to represent $F_{i, L}$ in the following paragraphs.
\begin{equation}
    F_{local}=H_{local}(F),
\end{equation}
\begin{equation} \label{eq:global}
    F_{global}=H_{global}(F). 
\end{equation}

We then detail the local and global branches. The local branch is CNN based, as shown in Figure \ref{fig4}. Instead of using a single-layer convolution, we insert a residual connection and two convolution layers to increase the expressiveness of the model. The extraction of $F_{local}$ can be also written as, 

\begin{equation}
    F_{local}=H_{conv}(F)+F
\end{equation}

\noindent where $H_{conv}$(·) denotes a simple block containing three layers. Specifically, it consists of two $3 \times 3$ convolution layers and a LeakyReLU layer. 

In the global branch, we use the spectrum transform structure in accordance with \cite{chi2020fast}. It can transform the conventional spatial features into the frequency domain to extract the global features by 2-D FFT and perform the inverse 2-D FFT operation to produce final spatial domain features for future feature fusion. The $H_{global}$ in Eq. \ref{eq:global} can also be re-written as, 

\begin{equation} 
    F' = \mathcal{C}(F)
\end{equation}
\begin{equation} 
    F_{frequency} =  \mathcal{C''}(H_{IFFT}( \mathcal{C'}(H_{FFT}(F')))+F')
\end{equation}

\noindent
where $H_{FFT}$(·) is the channel-wise 2-D FFT operation. $H_{IFFT}$(·) is the inverse 2-D FFT operation. $\mathcal{C}$, $\mathcal{C'}$ and $\mathcal{C''}$ denote the used three convolution layers in the global branch. 

After obtaining the features from both branches, we finally use a single $1\times1$ convolution layer $\mathcal{C}_f$ to fuse the two features and reduce the number of channels by half. 

\begin{equation} \tag{9}
    F_{FFB}=\mathcal{C}_f ([F_{local}, F_{global}])
\end{equation}

\noindent
where $[\cdot]$ is the concatenation operation.

\subsubsection{Cross-View Interaction. } \label{rcam}

\begin{figure}[hbtp]
\centering
\includegraphics[width=\linewidth]{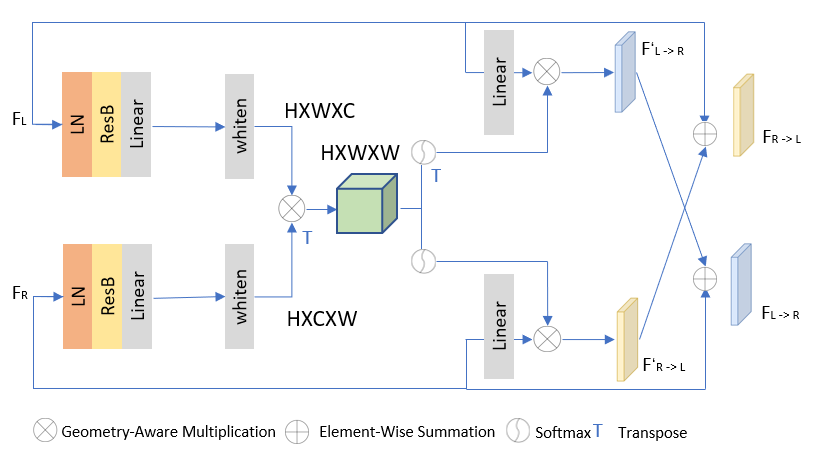}
\caption{Residual Cross Attention Module (RCAM).}
\label{fig5}
\end{figure}

In this section, we show the details of the proposed Residual Cross Attention Module (RCAM). The structure of RCAM is demonstrated in Figure \ref{fig5}. It is based on Scaled Dot Product Attention \cite{vaswani2017attention} and inspired by all the previous cross attention modules \cite{song2020stereoscopic, wang2019learning, wang2021symmetric, ying2020stereo}, which computes the dot products of the query with all keys and applies a softmax function to obtain the weights on the values: 

\begin{equation} \tag{10}
    Attention(Q, K, V) = softmax(QK^T/\sqrt{C})V
\end{equation}
\noindent where $Q \in R^{H \times W \times C}$ is a query matrix projected by the source intra-view feature (e.g., left-view), and $K, V \in R^{H \times W \times C}$ are key, value matrices projected by target intra-view feature (e.g., right-view). Here, H, W, and C represent the height, width and number of channels of the feature map. Since stereo images are highly symmetric under epipolar constraint \cite{wang2021symmetric}, we follow NAFSSR \cite{chu2022nafssr} to calculate the correlation of cross-view features along the W dimension.
In detail, given the input stereo intra-view features $F_L,F_R \in R^{H \times W \times C}$, we can get layer normalized stereo features $\bar{F_L}$= LN($F_L$) and $\bar{F_R}$= LN($F_R$). Next, a residual block (Resb) is applied to the process, and the processed feature is separately fed into two $1\times1$ convolutions and obtain $\hat{F_L}$ and $\hat{F_R}$. We then follow \cite{wang2021symmetric} to feed $\hat{F_L}$ and $\hat{F_R}$ to a whiten layer to acquire normalized features to establish disentangled pairwise parallax attention according to the following two equations:
\begin{equation} \tag{11}
    \bar{F_L}'(h, w, c) = \hat{F_L}(h, w, c) - \frac{1}{W}\sum_{i=1}^W\hat{F_L}(h,i,c)
\end{equation}
\begin{equation} \tag{12}
    \bar{F_R}'(h, w, c) = \hat{F_R}(h, w, c) - \frac{1}{W}\sum_{i=1}^W\hat{F_R}(h,i,c)
\end{equation}
\noindent Then a geometry-aware multiplication will be adopted between $bar{F_L}'$ and $\bar{F_R}'$:
\begin{equation} \tag{14}
    Attention = \bar{F_L}' \otimes \bar{F_R}'
\end{equation}
The bidirectional cross-attention between left-right views is calculated by: 
\begin{equation} \tag{15}
    F_{R->L} = Attention(W_{1}^L\bar{F_L}, W_{1}^R\bar{F_R}, W_{2}^RF_R),
\end{equation}
\begin{equation} \tag{16}
    F_{L->R} = Attention(W_{1}^R\bar{F_R}, W_{1}^L\bar{F_L}, W_{2}^LF_L),
\end{equation}
\noindent where $W_1^L , W_1^R , W_2^L$ and $W_2^R$ are projection matrices. Note that we can calculate the left-right attention matrix only once to generate both $F_{R->L}$ and $F_{L->R}$ (as shown in Figure \ref{fig5}). Finally, the interacted cross-view information $F_{R->L} , F_{L->R}$  and intra-view information $F_L, F_R$ are fused by element-wise addition same as NAFSSR \cite{chu2022nafssr}: 
\begin{equation} \tag{15}
    F_{L,out}= \gamma LF_{R->L} + F_L
\end{equation}
\begin{equation} \tag{15}
    F_{R,out} = \gamma RF_{L->R} + F_R
\end{equation}

\noindent where $\gamma$L and $\gamma$R are trainable channel-wise scales and initialized with zeros for stabilizing training.

\subsection{Training Strategies} \label{Strategies}
\noindent \textbf{Rectangular Training Patches.} In stereo image SR tasks, it is common to train models with small squared patches cropped from full-resolution images \cite{yang2020learning, wang2021symmetric}. Due to the fact that disparity of the stereo images existing along the epipolar line, some models use $30 \times 90$ rectangular patches to train the stereoSR models \cite{jin2022, zhang2022swinfir}. We empirically find that the patch size does affect the model performance and we show the experimental results  in Table \ref{tab:4}. These patches are randomly flipped horizontally and vertically for data augmentation. 

\noindent
\textbf{Dropout Rate and Stochastic Depth.} To further utilize the training data, we adopt stochastic depth \cite{huang2016deep} and dropout \cite{kong2022reflash} as regularization. The results of using different stochastic depth and dropout rates during model training can be found in Table \ref{tab:2}. 

\noindent
\textbf{Loss Functions.} 
We use the pixel-wise L1 distance between the SR and ground-truth stereo images in the NTIRE 2023 Stereo Image Super Resolution Challenge Track 1 \cite{Wang2023NTIRE}:
\begin{equation} 
    L_{SR}= ||I_L^{SR}-I_L^{HR}||_1 + ||I_R^{SR}-I_R^{HR}||_1,
\end{equation}
\noindent
where $I_L^{SR}$ and $I_R^{SR}$ are respectively the super-resolved left and right images. $I_L^{HR}$ and $I_R^{HR}$ are the ground truths. 

For the Challenge Track2, inspired by \cite{zhu2017unpaired, yu2021two}, we adopt a combination of perceptual loss and L1 loss to enhance supervision in the high-level feature space, as outlined below:
\begin{equation} 
    L_{Final} = L_{SR} + 0.01*L_{Per}
\end{equation}
\begin{equation}\label{eq:1}
    L_{Per} = \frac{1}{N}\sum_j\frac{1}{C_jH_jW_j}{||\phi_j(f_\theta(I^{LR})) - \phi_j(I^{HR})||_2^{2}}.
\end{equation}

The VGG-16 \cite{simonyan2014very}, pre-trained on ImageNet, serves as the loss network $\phi$. The loss function, expressed in equation \ref{eq:1}, uses the left and right low resolution input image $I_L^{LR}$, $I_R^{LR}$ and their correspondence high resolution ground truth images $I_L^{HR}$, $I_R^{HR}$. And the super-resolved images $I^{SR}$, generated by the SwinFSR model are denoted by $f_\theta(\cdot)$, where $\phi_j(\cdot)$ represents the feature map with a size of $C_j \times H_j \times W_j$. $j$ denote the $j$-th layer of VGG-16. Moreover, the L2 loss is utilized as the feature reconstruction loss and the perceptual loss function employs N features.
\section{Experiments}
\subsection{Implementation Details}
\noindent\textbf{Evaluation Metrics.} The evaluation metrics used are peak signal-to-noise ratio (PSNR) and structural similarity (SSIM). These metrics are calculated in the RGB colour space using a collection of stereo images obtained by averaging the left and right views. Table \ref{modelsize} displays the influence of varying architecture, including three different sizes of SwinFSR by modifying the number of blocks. These networks are identified as SwinFSR-S (Small), SwinFSR-B (Big), and SwinFSR-L (Large). \\
\textbf{Training Detail.} All models are optimized by the Adam \cite{kingma2014adam} with $\beta1$ = 0.9 and $\beta2$ = 0.9. The learning rate is set to  $1e^-4$ and decreased to $1e^-5$ with a cosine annealing strategy \cite{loshchilov2016sgdr}. If not specified, models are trained on $30 \times 90$ patches with a batch size of 1 for $7e^6$ iterations. The window size of the model is $6 \times 15$. Data augmentation includes horizontal/vertical flips and RGB channel shuffle are used. \\
\textbf{Datasets.} To conduct our experiments, we utilize the training and validation datasets provided by the NTIRE Stereo Image SR Challenge \cite{wang2022ntire}. Specifically, we use 800 stereo images from the training set of the Flickr1024 \cite{wang2019flickr1024} dataset as our training data and 112 stereo images from the validation set of the same dataset as our validation set. The low-resolution images are created by downsampling using the bicubic method. In addition, we follow the dataset splits in \cite{chu2022nafssr} to conduct a comparison on KITTI 2012 \cite{geiger2012we}, KITTI 2015 \cite{menze2015object}, Middlebury \cite{scharstein2014high} and Flickr1024 \cite{wang2019flickr1024}. 

\begin{table}[t]
\centering
\caption{The performance of different SwinFSRs in size. }
\scalebox{1}{
\setlength{\tabcolsep}{3.4mm}{
\begin{tabular}{|c | c c c |}
     \toprule
     Model & \#RSFTBs & \#Params & PSNR  \\ 
     \hline
     SwinFSR-S & 4 & 9.76M & 23.8319  \\ 
   
     SwinFSR-B & 6 & 14.01M & 23.9630  \\
     
     \textbf{SwinFSR-L} & \textbf{12} & \textbf{26.75M} & \textbf{24.1940}  \\
    \bottomrule        
    \end{tabular}}}
\label{modelsize}
\end{table}

\subsection{Ablation Study}

\begin{table}[t]
\centering
\caption{
The influence of different cross-attention modules. We here report the results in both PSNR and SSIM for $4 \times $SR. TTA represents the test-time augmentation. SwinFSR-L is used to conduct this analysis.
}
\scalebox{0.88}{
\begin{tabular}{|c| c c c c |} 
 \toprule
 \multicolumn{1}{|c|}{\multirow{2}{*}{\textbf{Modules}}}  & \multicolumn{2}{c}{PSNR}  & \multicolumn{2}{c|}{SSIM}  \\ 
 &  w/o TTA & w. TTA & w/o TTA & w. TTA  \\ 
     \midrule
 - & 23.6921 & 23.7714 & 0.7380 & 0.7397 \\
 
 \hline
  biPAM\cite{wang2021symmetric} & 23.8883 & 24.0510 & 0.7432 & 0.7520 \\
 \hline
  SAM\cite{ying2020stereo} & 22.3834 & 22.4366 & 0.6690 & 0.6715  \\
 \hline
  SCAM\cite{chu2022nafssr} & 24.0882 & 24.1926 & 0.7564 & \textbf{0.7616} \\
 \hline
  \textbf{RCAM} & \textbf{24.1233} & \textbf{24.1940} & \textbf{0.7583} & 0.7598 \\
 \bottomrule
\end{tabular}}      

\label{tab:t}
\end{table}

\begin{table}[t]
\centering
\caption{The efficiency comparison between several cross-attention modules. We replace the cross-attention module in  SwinFSR-L to conduct the analysis. Training time is the cost for $4 \times $SR on Flickr1024 \cite{wang2019flickr1024} training set.}
\scalebox{1.0}{
\begin{tabular}{|c| c  c c|} 
 \toprule
 Modules & Params & Time/Epoch & Speedup\\  
 \hline
 SAM\cite{ying2020stereo} & 32.72M  & 1259ms & -\\
 \hline
 SCAM\cite{chu2022nafssr} & 25.00M  & 988ms & \%21.5\\
 \hline
 RCAM & 26.75M  & 1065ms & \%15.4\\
 \bottomrule
\end{tabular}}           
\label{tab:t2}
\end{table}

\begin{table*}[t]
\centering
\caption{
The influence of different dropout rates. We here report the results in both PSNR and SSIM for $4 \times $SR. TTA represents the test-time augmentation. SwinFSR-S is used to conduct this analysis.}
\scalebox{0.9}{
\setlength{\tabcolsep}{5mm}{
\begin{tabular}{|c| c c c c c|}
     \toprule
     \multicolumn{1}{|c|}{\multirow{2}{*}{\textbf{Model}}} & \multicolumn{1}{c}{\multirow{2}{*}{\textbf{Dropout Rate}}} & \multicolumn{2}{c}{PSNR}  & \multicolumn{2}{c|}{SSIM} \\
      &  & w/o TTA & w. TTA & w/o TTA & w. TTA  \\ 
     \midrule
     
     \multicolumn{1}{|c|}{\multirow{4}{*}{\textbf{SwinFSR-S}}} & N/A & 23.7304 & 23.8191 & 0.7430 & 0.7451 \\ 
     \cline{2-6}
     & \textbf{0.1} & \textbf{23.8319} & \textbf{23.9240} & \textbf{0.7471} & \textbf{0.7492}  \\
     \cline{2-6}
     & 0.3 & 23.8319 & 23.9230 & 0.7470 & 0.7491 \\
     \cline{2-6}
     & 0.5 & 21.6377 & 22.4352 & 0.6365 & 0.6767 \\
     \midrule                 
    \end{tabular}}}
\label{tab:1}
\end{table*}

\noindent \textbf{Residual Cross-Attention Modules.} Here, all the experiments are conducted using SwinFSR-L. To show the effectiveness of RCAM,  we substitute the cross-attention module in  SwinFSR-L with several SOTA approaches, such as biPAM \cite{wang2021symmetric}, SAM \cite{ying2020stereo}, SCAM \cite{chu2022nafssr} and baseline (without cross-attention module.).  Table \ref{tab:t} shows the 4 $\times $ SR results   on Flickr1024 \cite{wang2019flickr1024}. First, 
when compared with the baseline that only explored intra-view information, our method is 0.4 dB higher than the baseline in PSNR.
Furthermore, compared with biPAM, SCAM, and SAM, our RCAM achieves improvements of 0.235 dB, 0.035 dB, and 1.740 dB, respectively.

In addition, to further show the efficiency of our RCAM, we provide in Table \ref{tab:t2} by the number of parameters and training time. 
It can be observed that our proposed RCAM has fewer parameters and training time than that of SAM. It is worth mentioning that both SCAM and our RCAM do not handle occlusion problems when performing cross-view integration. Interestingly, we find using SCAM and RCAM does not jeopardize the performance but can help achieve better PSNR and faster training.
These outcomes emphasize the importance of a well-designed cross-attention model and the critical impact of integrating both cross-view information  and intra-view information. 


\noindent\textbf{Test Time Augmentations.} Although test-time augmentation (TTA) has been commonly utilized in competitions to enhance performance, its usefulness in stereo SR tasks has not been proven. Here, we use horizontal and vertical flips as our TTA strategy. To evaluate the effectiveness of TTA in this task, we assess each model's inference results using the NTIRE 2023 Stereo Image SR validation dataset \cite{Wang2023NTIRE}. The results, presented in Table \ref{tab:t}, \ref{tab:1}, \ref{tab:4} and \ref{tab:2}, demonstrate that employing TTA is always beneficial. This phenomenon suggests that TTA is indeed effective for stereo SR tasks. 

\noindent\textbf{Dropout.} According to \cite{kong2022reflash}, adding only one line of dropout layer can significantly improve the model performance. We thus follow \cite{kong2022reflash} to put the dropout layer before the last convolution layer. Then, we use SwinFSR-S to investigate the impact of the dropout rate during training. In Table \ref{tab:1}, we report results on Flickr1024 \cite{wang2019flickr1024} validation set. Compare to the SwinFSR-S model without the specific dropout layer, with a 10\%  dropout rate, the PSNR result can be improved by 0.102 dB. However, when we increase the dropout rate to 30\%, the performance does not change.  When it comes to 50\%, half of the nodes are dropped during the training, which makes the performance decrease by 2.194 dB. 

\begin{table}[t]
\centering
\caption{The influence of different window sizes and training patch sizes. We here report the results in both PSNR and SSIM for $4 \times $SR. TTA represents the test-time augmentation. SwinFSR-S is used to conduct this analysis.}
\scalebox{0.9}{
\setlength{\tabcolsep}{1mm}{
\begin{tabular}{|c c c c c c |} 
 \toprule
 Patch & Window & PSNR & PSNR w. TTA & SSIM & SSIM w. TTA \\
 \hline
 $32 \times 32$ & 4$\times$4 & 23.52 & 23.63 & 0.734 & 0.738  \\
 \hline
 $32 \times 32$ & 8$\times$8 & 23.57 & 23.65 & 0.734 & 0.737 \\
 \hline
 $30 \times 90$ & 3$\times$9 & 23.65 & 23.74 & 0.739 & 0.741 \\
 \hline
 \textbf{$30 \times 90$} & \textbf{6$\times$15} & \textbf{23.83} & \textbf{23.92} & \textbf{0.747} & \textbf{0.749} \\
 \bottomrule             
\end{tabular}}}

\label{tab:4}
\end{table}

\noindent\textbf{Window Size and Training Patch Size.} According to \cite{zhang2022swinfir}, a larger window size can enhance the performance of stereoSR. Here, we use SwinFSR-S to further investigate the impact of window size. Table \ref{tab:4} reports results on Flickr1024 \cite{wang2019flickr1024} test set. First, while using the same squared training patch size, a larger window size will improve the performance of SwinFSR-S by 0.049 dB. If further changing the training patch sizes to be rectangular according to the epipolar stereo disparity \cite{wang2021symmetric}, the performance will be increased by 0.087 dB. Moreover, increasing window size while using rectangular training patches boost the performance by 0.178 dB. Due to the limitation of the GPU resources, we do not further enlarge the window size and training patch size. This shows that the rectangular training patch and larger local window size indeed can help improve the feature extraction ability across stereo images.

\begin{table*}[t]
\centering
\caption{
The influence of stochastic depth. We here report the results in both PSNR and SSIM for $4 \times $SR. TTA represents the test-time augmentation. SwinFSR-L is used to conduct this analysis.}
\scalebox{0.85}{
\setlength{\tabcolsep}{5mm}{
\begin{tabular}{|c| c c c c c |}
     \toprule
     \multicolumn{1}{|c|}{\multirow{2}{*}{\textbf{Model}}} & \multicolumn{1}{c}{\multirow{2}{*}{\textbf{Stochastic Depth}}} & \multicolumn{2}{c}{PSNR}  & \multicolumn{2}{c|}{SSIM} \\
      &  & w/o TTA & w. TTA & w/o TTA & w. TTA \\ 
     \midrule
     \multicolumn{1}{|c|}{\multirow{4}{*}{\textbf{SwinFSR-L}}} & N/A & 23.9516 & 24.0442 & 0.7518 & 0.7537 \\
     \cline{2-6}
      & 0.1 & 24.0786 & 24.1679 & \textbf{0.7573} & \textbf{0.7591}  \\
     \cline{2-6}
      & \textbf{0.2} & \textbf{24.0928} & \textbf{24.1773} & 0.7470 & 0.7491 \\
     \cline{2-6}
     & 0.3 & 23.9719 & 24.1035 & 0.7518 & 0.7548  \\
     \bottomrule
    \end{tabular}}}
 
\label{tab:2}
\end{table*}

\noindent\textbf{Stochastic Depth.} As per the research conducted by \cite{chu2022nafssr}, a deeper stochastic depth can improve the performance of stereoSR. Therefore, we employ SwinFSR-L to examine how stochastic depth affects our Swin Transformer based model. Our results based on the validation set of Flickr1024 \cite{wang2019flickr1024} are presented in Table \ref{tab:2}. During training, incorporating 10\% stochastic depth \cite{huang2016deep} lead to a 0.102 dB improvement in PSNR. When using 20\% stochastic depth, the performance of SwinFSR-L improves slightly by 0.1014 dB. However, setting the stochastic depth to 30\% results in a performance decrease of 0.121 dB, but it still outperforms the baseline. This suggests that larger models have a tendency to overfit the Flickr1024 training data. However, incorporating stochastic depth can help enhance the overall performance and generalization ability of the networks.

\begin{table*}[t]
\centering
\caption{Comparison with several state-of-the art methods for $4 \times $SR  on the KITTI 2012 \cite{geiger2012we}, KITTI 2015 \cite{menze2015object}, Middlebury \cite{scharstein2014high} and Flickr1024 \cite{wang2019flickr1024} datasets. The number of parameters  is denoted by "Params". Numbers reported for each dataset are in PSNR/SSIM. } 
\scalebox{0.9}{
\setlength{\tabcolsep}{5mm}{
\begin{tabular}{|c | c | c | c| c|c|} 
     \toprule
     Model & \#Params & KITTI2012 & KITTI2015 & Middlebury & Flickr1024\\ 
     \hline
     VDSR & 0.66M & 25.60/0.7722 & 25.32/0.7703 & 27.69/0.7941 & 22.46/0.6718\\
     \hline
     EDSR & 38.9M & 26.35/0.8015 & 26.04/0.8039 & 29.23/0.8397 & 23.46/0.7285\\
     \hline
     RDN & 22.0M & 26.32/0.8014 & 26.04/0.8043 & 29.27/0.8404 & 23.47/0.7295\\
     \hline
     RCAN & 15.4M & 26.44/0.8029 & 26.22/0.8068 & 29.30/0.8397 & 23.48/0.7286\\
     \hline
     StereoSR & 1.42M & 24.53/0.7555 & 24.21/0.7511 & 27.64/0.8022 & 21.70/0.6460\\
     \hline
     SRRes+SAM & 1.73M & 26.44/0.8018 & 26.22/0.8054 & 28.83/0.8290 & 23.27/0.7233\\
     \hline
     PASSRnet & 1.42M & 26.34/0.7981 & 26.08/0.8002 & 28.72/0.8236 & 23.31/0.7195\\
     \hline
     iPASSR & 1.42M & 26.56/0.8053 & 26.32/0.8084 & 29.16/0.8367 & 23.44/0.7287\\
     \hline    
     SSRDE-FNet & 2.24M & 26.70/0.8082 & 26.43/0.8118 & 29.38/0.8411 & 23.59/0.7352\\
     \hline
     SwiniPASSR-M2 & 22.81M & -/- & -/- & -/- & 24.13/0.7579\\
     \hline
     NAFSSR-L & 23.83M & 27.12/0.8194 & 26.96/0.8257 & 30.20/0.8605 & 24.17/0.7589\\ 
     \midrule\midrule
     SwinFSR-S (ours) & 9.76M & 27.03/0.8143 & 26.83/0.8213 & 32.45/0.8891 & 23.83/0.7471\\
     \hline
     SwinFSR-B (ours) & 14.01M & 27.07/0.8151 & 26.87/0.8222 & 32.69/0.8910 & 23.96/0.7510\\
     \hline
     \textbf{SwinFSR-L (ours)} & 26.75M & \textbf{27.24/0.8195} & \textbf{27.00/0.8257} & \textbf{32.73/0.8915} & \textbf{24.19/0.7598}\\
     \bottomrule
    \end{tabular}}}
 
\label{tab:all}
\end{table*}

\subsection{Comparison with the state-of-the-art methods}

To make a fair comparison with previous works, we follow  the dataset splits in NAFSSR \cite{chu2022nafssr} to train and test our method on four representative datasets, i.e., KITTI 2012 \cite{geiger2012we}, KITTI 2015 \cite{menze2015object}, Middlebury \cite{scharstein2014high} and Flickr1024 \cite{wang2019flickr1024}. Specifically, we generate low-resolution images by applying bicubic downsampling to high-resolution (HR) images with a scaling factor of 4. Then we randomly crop  $30\times 90$ patches from stereo images  as inputs. 
During training, we set all the hyperparameters to the best possible ones given by our ablation studies, such as dropout rate, window size, and stochastic depth.
Additionally, we 
employ horizontal and vertical flips as our test-time augmentation. For the results on Flickr1024, we perform results ensemble by collecting the top three performed models on the validation set and averaging their inference results on the test set as the final results (the same strategy we used in the NTIRE2023 challenge \cite{Wang2023NTIRE}). For the other three datasets, we report the best performance without an ensemble. 

Table \ref{tab:all} presents the quantitative comparison of SwinFSR and several SOTA super-resolution methods. Our comparison includes single SR methods such as VDSR \cite{kim2015accurate}, EDSR \cite{lim2017enhanced}, RDN \cite{zhang2018residual}, RCAN \cite{zhang2018image}, and SwinIR \cite{liang2021swinir}, as well as stereo SR methods including StereoSR \cite{jeon2018enhancing}, PASSRnet \cite{wang2019learning},  SRRes+SAM \cite{ying2020stereo}, iPASSR \cite{wang2021symmetric}, SRRDE-FNet \cite{dai2021feedback}, SwiniPASSR \cite{jin2022}, and NAFSSR \cite{chu2022nafssr}. The evaluation metrics used are PSNR and SSIM, and the dataset used for testing are KITTI 2012 \cite{geiger2012we}, KITTI 2015 \cite{menze2015object}, Middlebury \cite{scharstein2014high} and Flickr1024 \cite{wang2019flickr1024}. By checking throughout the table, it can be observed that our method outperforms all the compared approaches on the four datasets.  These results further validate the effectiveness of our proposed method.

\subsection{NTIRE Stereo Image SR Challenge} 
\noindent We submit a result obtained by the presented approach to the NTIRE 2023 Stereo Image Super-Resolution Challenge Track 1 and 2 \cite{Wang2023NTIRE}. In order to maximize the potential performance of our method, we adopt the stochastic depth \cite{huang2016deep} with 0.2 probability to improve the model's generality ability. 
During test time, we adopt horizontal and vertical flips as our TTA strategy. Finally, we average the SR images from the top 3 performance models on the validation set for our final submission. As a result, our final submission achieves 24.1940 dB in PSNR on the validation set and won a ninth place with 23.7121 dB in PSNR on the test set.

\section{Conclusion}
\noindent The goal of this paper is to introduce a novel network called SwinFSR for enhancing the resolution of stereo images. To achieve this, we utilize a series of RSFTBlocks to extract intra-view features with enlarged reception fields and propose residual stereo cross-attention modules (RCAMs) to interact between both intra-view and cross-view features. Additionally, we explore the best possible hyperparameters, such as dropout rate, training patch size, window size, and stochastic depth and found the best values are 10\%,  $30 \times 90$, $6 \times 15$ and 20\% respectively. Extensive ablation studies demonstrate the effectiveness of the proposed method.

{\small
\bibliographystyle{ieee_fullname}
\bibliography{egbib}
}

\end{document}